\documentclass[runningheads]{llncs}
\usepackage{amsmath,graphicx}
\usepackage{xcolor}
\usepackage{comment}
\usepackage{subcaption}
\usepackage{longtable}
\usepackage{url}
\usepackage{amssymb}
\usepackage{array}
\usepackage{booktabs}
\usepackage{subcaption}
\usepackage{adjustbox, makecell}
\usepackage[T1]{fontenc}
\usepackage{graphicx}
\begin{document}
\title{Saudi Sign Language Translation Using T5}
\author{Ali Alhejab\inst{1}\orcidID{0009-0005-3090-1453} \and
Tomáš Železný \inst{2}\orcidID{0000-0002-0974-7069} \and
Lamya Alkanhal \inst{3}\orcidID{0009-0006-3658-6738} \and
Ivan Gruber\inst{2}\orcidID{0000-0003-2333-433X} \and
Yazeed Alharbi\inst{1}\orcidID{0000-0002-8073-1959} \and
Jakub Straka\inst{2}\orcidID{0000-0002-9981-1326} \and
Václav Javorek\inst{2}\orcidID{0009-0008-1019-8854} \and
Marek Hrúz\inst{2}\orcidID{0000-0002-7851-9879} \and
Badriah Alkalifah\inst{1} \and
Ahmed Ali \inst{1}\orcidID{0000-0003-2506-5647}
}
\authorrunning{Ali Alhejab et al.}

\institute{HUMAIN, Riyadh, Saudi Arabia \and 
Department of Cybernetics and New Technologies for the Information Society, University of West Bohemia, Pilsen, Czech Republic \and 
Saudi Data \& AI Authority , Riyadh, Saudi Arabia \\
\email{aalhejab@humain.ai, zeleznyt@ntis.zcu.cz, yaharbi@humain.ai, grubiv@ntis.zcu.cz}
}

%  Technická 8, 301 00 Plzeň
\maketitle              % typeset the header of the contribution
\begin{abstract}
This paper explores the application of T5 models for Saudi Sign Language (SSL) translation using a novel dataset. The SSL dataset includes three challenging testing protocols, enabling comprehensive evaluation across different scenarios. Additionally, it captures unique SSL characteristics, such as face coverings, which pose challenges for sign recognition and translation. In our experiments, we investigate the impact of pre-training on American Sign Language (ASL) data by comparing T5 models pre-trained on the YouTubeASL dataset with models trained directly on the SSL dataset. Experimental results demonstrate that pre-training on YouTubeASL significantly improves models' performance (roughly $3\times$ in BLEU-4), indicating cross-linguistic transferability in sign language models. Our findings highlight the benefits of leveraging large-scale ASL data to improve SSL translation and provide insights into the development of more effective sign language translation systems. Our code is publicly available at our GitHub repository\footnote{\url{https://github.com/signforall/t5-training-scripts}}.

\end{abstract}
\section{Introduction}
Sign languages (SLs) are rich, fully developed natural languages that serve as the primary means of communication for Deaf communities worldwide. Unlike spoken languages, SLs utilize visual-gestural modalities: hand shapes, movements, facial expressions, and body language, to convey meaning. According to recent estimates, more than 70 million Deaf people use SL, and there are over 300 distinct SLs in use globally, reflecting the diverse cultural and linguistic heritage of Deaf communities\footnote{\url{https://www.handtalk.me/en/blog/nteresting-facts-about-sign-languages/}}.

In many countries, there is increasing recognition of the need for inclusive communication in public and private institutions such as banks, hospitals, and schools. Effective communication between the Deaf community and these institutions is essential for ensuring equitable access to critical services. For instance, in healthcare settings, the presence of qualified SL interpreters has been shown to significantly improve patient understanding and satisfaction, while in educational environments, SL is the preferred language for many Deaf students to learn complex concepts in their native tongue (see NAD Position Statement on Health Care Access for Deaf Patients, 2020\footnote{\url{https://shorturl.at/tQ1De}}). Similarly, financial institutions and government agencies are progressively adopting SL interpretation services to better serve Deaf clients, highlighting the importance of culturally and linguistically appropriate communication.

%One of the main challenges in sign language translation (SLT) is the scarcity of %training data, especially for sign languages that are not well represented in %public sources. As a result, recent research has begun to explore SLT in a %multilingual context by leveraging corpora that encompass data from multiple %SLs~\cite{YouTube-SL-25}. These approaches suggest that incorporating %multilingual corpora not only mitigates data scarcity issues but also allows %models to exploit shared linguistic structures among different SLs, leading to %improved translation quality.

% SSL

One of the main challenges in sign language translation (SLT) is the scarcity of training data, particularly for sign languages that are underrepresented in publicly available resources. Recent research has explored SLT in a multilingual context by leveraging corpora from multiple sign languages (SLs), which not only helps address the data scarcity issue but also allows models to exploit shared linguistic structures, leading to improved translation quality \cite{YouTube-SL-25}. This challenge is especially pronounced for under-resourced sign languages like Saudi Sign Language (SSL), primarily used by the Deaf community in Saudi Arabia. SSL is characterized by unique region-specific gestures, non-manual markers (such as facial expressions and body movements), and syntactic structures that reflect both cultural influences and elements of spoken Arabic. Unlike Unified Arabic Sign Language, a standardized system used across many Arab countries, SSL has developed independently, resulting in distinct grammar and vocabulary. For instance, \cite{altamimi2023argument} demonstrates that SSL follows unconventional sentence structures and word orders, differing from the broader Arabic SL standard, while \cite{sprenger2012observations} highlights unique non-manual markers and syntactic patterns that further differentiate SSL from other Arabic SL variants.

The main contributions of this paper are as follows: 
Firstly, we propose a processing pipeline directly tailored for sign language videos.
Secondly, we demonstrate the effectiveness of pre-training on a different SL to improve generalization performance. We explore this idea by applying it to Saudi Sign Language (SSL), leveraging the ASL dataset YouTubeASL~\cite{yasl} to pre-train a T5-based SLT model. We then compare the results with a model trained from scratch. Using a pose-based approach that omits the appearance of signers, our findings show that cross-lingual pre-training significantly enhances performance, highlighting its potential for low-resource SLs. 

\section{Related Work}
Sign Language Translation has advanced through both gloss-based approaches~\cite{zhou2021improving,chen2022simple,Spoter}, which use glosses - structured linguistic representations of signs - for improved alignment, and gloss-free approaches~\cite{yin2023gloss,zhou2023gloss,SignBERT+}, which aim to learn direct mappings from visual features to text. While gloss-based methods benefit from explicit supervision, recent gloss-free approaches have become increasingly popular by utilizing multimodal learning techniques. The advancement of Large Language Models (LLMs) has further improved gloss-free SLT, as seen in~\cite{Sign2GPT,SSVP-SLT,LLaVASLT}, by the use of better pre-trained textual representations to improve translation accuracy. 

%The scarcity of annotated data continues to be a major challenge, especially for %under-resourced SLs such as Saudi Sign Language. SSL, primarily used by the Deaf %community in Saudi Arabia, is characterized by its unique set of region-specific %gestures, non-manual markers (such as facial expressions and body movements), and %syntactic structures that reflect both cultural influences and elements of spoken %Arabic. Unlike the Unified Arabic Sign Language, which is standardized system used %across many Arab countries, SSL has developed independently and naturally. This %has resulted in unique grammar and vocabulary. For example, %in~\cite{altamimi2023argument} the authors demonstrate that SSL follows %unconventional sentence structures and word order, that differ from those found %in the broader Arabic SL standard. Similarly, \cite{sprenger2012observations}
%~highlights unique non-manual markers and syntactic patterns that differentiate %SSL from its Arabic counterparts.

% Bilingual transfer methods
While studies about SSL like~\cite{s24103112} have focused on recognition rather than full translation, multilingual corpora such as~\cite{JWSign} have shown the potential for cross-lingual adaptation. Bilingual transfer methods, like the ones used in~\cite{ASLtoISL}, show a high added value when using high-resource SLs to improve translations for lower-resource ones.

% Data
Several large-scale datasets are being used for SLT training~\cite{YouTube-SL-25,yasl,PHOENIX-2014T,how2sign}. However, privacy concerns and high annotation costs limit their scalability. In~\cite{SSVP-SLT}, the authors take this issue on by introducing self-supervised pre-training on anonymized videos, and~\cite{LLaVASLT} uses hierarchical visual encoders and multimodal tuning to find better sign language representations without gloss supervision.

% T5 justification
Transformer architectures, such as the Text-to-Text Transfer Transformer T5~\cite{T5}, have demonstrated significant effectiveness in SLT due to their encoder-decoder structure and multilingual capabilities. Studies~\cite{llmsgoodsignlanguagetranslators,yano2024multilingual} have demonstrated T5’s adaptability to multimodal input. Our work builds on this, using T5 as an SLT baseline while addressing the data limitations of SSL by employing multilingual transfer learning.

\begin{table*}[t]
    \centering
    \small
    \renewcommand{\arraystretch}{1.0}
    \setlength{\tabcolsep}{6pt}
    \begin{adjustbox}{width=\textwidth}
    \begin{tabular}{|c|c|c|c|c|c|c|c|}
        \hline
        \textbf{Split} & \textbf{Sents} & \textbf{Min} & \makecell{\textbf{Seen}\\\textbf{Sents}} & \makecell{\textbf{Seen}\\\textbf{Signers}} & \makecell{\textbf{\#}\\\textbf{Samples}} & \makecell{\textbf{\#}\\\textbf{Signers}} & \textbf{Gender} \\
        \hline
        Train & 24,111 & 2,017.82 & \checkmark & \checkmark & 1,900 & 16 & 4F, 12M \\
        Test 1 & 200 & 16.65 & x & x & 100 & 2 & 1F, 1M \\
        Test 2 & 1,297 & 107.95 & x & \checkmark & 100 & 11 & 3F, 10M \\
        Test 3 & 3,783 & 337.33 & \checkmark & x & 1,900 & 2 & 1F, 1M \\
        \hline
    \end{tabular}
    \end{adjustbox}
    \vspace{5pt}
    \caption{Dataset splits with details on number of sentences (Sents), minutes (Min), seen sentences/signers, etc.}
    \label{tab: data split}
\end{table*}

% \begin{table*}[t]
%     \centering
%     \renewcommand{\arraystretch}{1.0}
%     \setlength{\tabcolsep}{8pt} % Adjust column spacing for better alignment
%     \begin{tabular}{|c|c|c|c|c|c|c|c|}
%         \hline
%         \textbf{Split} & \textbf{Sentences} & \textbf{Minutes} & \textbf{Seen Sentences} & \textbf{Seen Signers} & \textbf{\# Samples} & \textbf{\# Signers} & \textbf{Gender} \\
%         \hline
%         Train & 24,111 & 2,017.82 & \checkmark & \checkmark & 1,900 & 16 & 4F, 12M \\
%         Test 1 & 200 & 16.65 & x & x & 100 & 2 & 1F, 1M \\
%         Test 2 & 1,297 & 107.95 & x & \checkmark & 100 & 11 & 3F, 10M \\
%         Test 3 & 3,783 & 337.33 & \checkmark & x & 1,900 & 2 & 1F, 1M \\
%         \hline
%     \end{tabular}
%     \caption{The SSL dataset splits with data details for sentences, minutes, seen sentences, seen signers, and other parameters.}
%     \label{tab: data split}
% \end{table*}

\section{Data}  
The dataset used in this study is the Saudi Sign Language corpus, which belongs to under-resourced SLs. In comparison with datasets such as YouTubeASL~\cite{yasl}, the number of recorded hours is significantly lower. However, one key advantage of SSL is that it enables a thorough assessment of model generalization.

\subsection{Dataset Composition}  
\begin{figure}[ht]  
    \centering
 \includegraphics[scale=0.53]{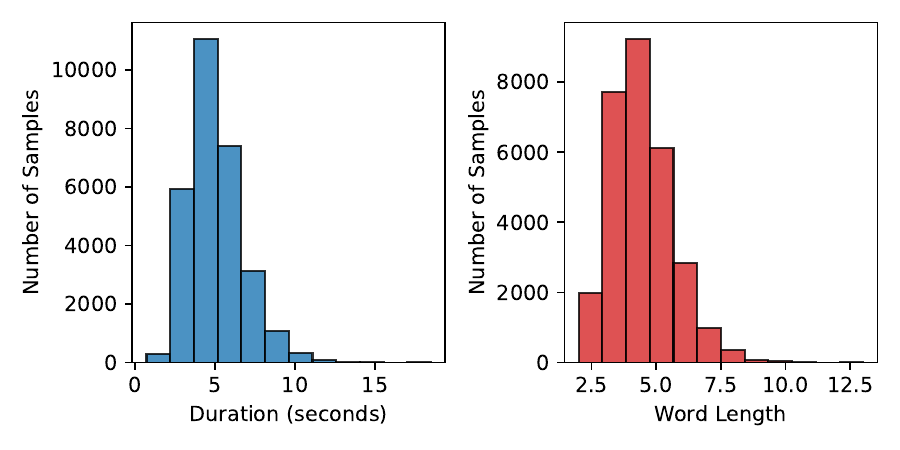} 
    \caption{Histogram of Signing Duration and Word Length.} % Add a caption
\label{fig:vide_stats_duration} % Add a label for referencing
\end{figure}

The SSL dataset comprises 2,000 unique sentences,   representing common expressions in the deaf community, spanning everyday communication and specialized domains (banking, law, education, healthcare, emergency services, and transportation). The original sentences are in Arabic and were translated into English for our experiments, as T5 and T5 v1.1 only support English. For mT5, which is multilingual, both the original Arabic sentences and the translated English sentences were used. The temporal distribution of the data is illustrated in Fig. \ref{fig:vide_stats_duration}, which highlights variations in sentence length and signing duration.

Eighteen signers participated in the recording, with 13~males and 5~females, resulting in a higher male representation. Notably, female signers’ faces were intentionally obscured (e.g., with masks or veils), while male signers’ faces remained visible. 

\subsection{Data Splits}  
The dataset is divided into train and multiple test splits (Table~\ref{tab: data split}). Each test scenario is designed to assess the model's ability to handle varying degrees of data exposure and distribution shifts. The test scenarios are as follows:

\textbf{Test 1:} This test is designed to assess the model's generalization ability to entirely unseen data. While the unseen sentences are unseen as a whole, they are composed of words seen during training. It evaluates the model's performance on both unseen sentences and signers, providing a measure of its robustness to novel input during inference.

\textbf{Test 2:} In this scenario, the model is tasked with generating translations for unseen sentences, but with signers it has encountered during training. This evaluates the model's ability to generalize to new, out-of-distribution sentences while leveraging prior knowledge of familiar signers and their signing styles.

\textbf{Test 3:} This test examines the model's performance on sentences it has been exposed to during training, but they are performed by unseen signers. This serves to gauge how well the model can generalize to the variety of ways people sign the same words.

The split highlights the key advantages of the dataset, as it provides more insight into model performance and generalization. Contrary to prior works, the splits enable direct measurement of whether the model is sensitive to signer appearance (Test 1 and Test 3) and word order (Test 1 and Test 2).

%Each signer was instructed to perform all 2,000 sentences. 
%However, the final usable dataset contains only 29,391 sentences (81.6\% of the expected 36,000), due to video corruption and incomplete recording from few signers. #

\subsection{Key Challenges and Limitations}  
Besides the challenging test splits, the SSL dataset presents a few challenges; some of them specific to SSL, and some of them are more general. The challenges in the data are as follows: 
\begin{itemize}  
    \item \textbf{Face Occlusion:} The covering of female faces may limit the model’s ability to learn important features from the lips or signs that rely on facial expression, such as question marks.
    \item  \textbf{Gender Imbalance:} The dataset includes SL data from 18 signers, with a notable gender imbalance (more male signers than female)
    \item  \textbf{Unbalanced Data Across Domains:} While the dataset spans multiple domains, the distribution of sentences across these domains may not be uniform.
\end{itemize}

\section{Methods}

\subsection{Video preprocessing}
SL datasets vary in recording conditions. Some, like How2Sign \cite{how2sign}, are captured in controlled environments with a single signer centered in the frame. Others, such as YouTubeASL~\cite{yasl}, contain videos recorded in the wild, where signers may appear at different distances from the camera, in varying positions, or alongside multiple people.

To standardize the data, we first preprocess the videos to ensure that the signing individuals are centered in the frame, have a normalized size, and that all videos have the same resolution across the dataset. Additionally, we extract pose features during this step.

Our preprocessing pipeline consists of multiple steps. First, we use lightweight YOLOv8-nano~\cite{Jocher_Ultralytics_YOLO_2023} to detect the rough body pose of all individuals in the frame. To simplify processing, we discard videos with multiple people, as tracking multiple individuals and identifying the signer throughout the video introduces complexity and potential misalignment between signing and translations. This step is utilized for the YouTubeASL dataset.

Next, we define the signing space, which in SL linguistics refers to the area where signing occurs. Inspired by~\cite{Spoter}, we represent this as a box centered between the shoulders, with a height and width four times the shoulder distance. If body pose keypoints fall outside this box, we expand it to include them. To create a stable bounding box for the entire video, we compute the signing space for each frame and take the median of the coordinates; this mitigates fluctuations caused by detection errors.

We then refine the signer’s pose using  MediaPipe~\cite{lugaresi2019mediapipe}, a more precise model for body, hand, and facial keypoints. MediaPipe performs better when the signer is centered in the frame, which especially benefits face and hand detection. Using the updated body keypoints, we adjust the signing space. In some cases, it is also necessary to determine the handedness of detected hand keypoints based on the Euclidean distance between wrist keypoints from the body pose and hand pose predictions.

\begin{figure}[htbp]
  \centering
  \begin{subfigure}[m]{0.48\textwidth}
    \centering
    \begin{subfigure}[m]{0.53\textwidth}
      \includegraphics[width=\linewidth]{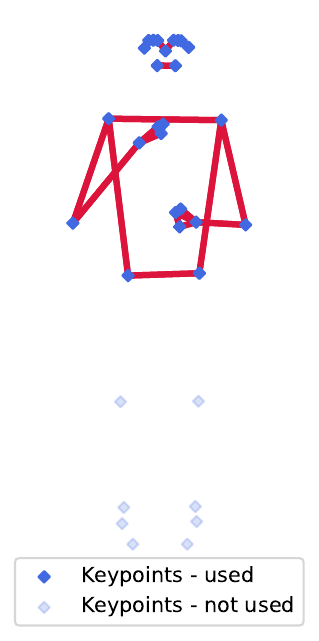}
    \end{subfigure}%
    \begin{subfigure}[m]{0.43\textwidth}
      \includegraphics[width=\linewidth]{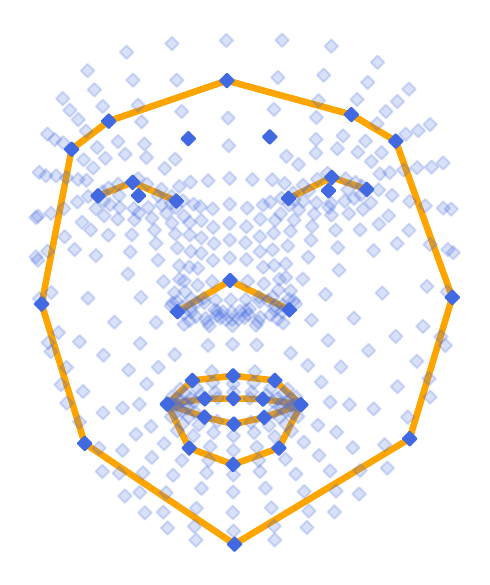}\\
      \includegraphics[width=\linewidth]{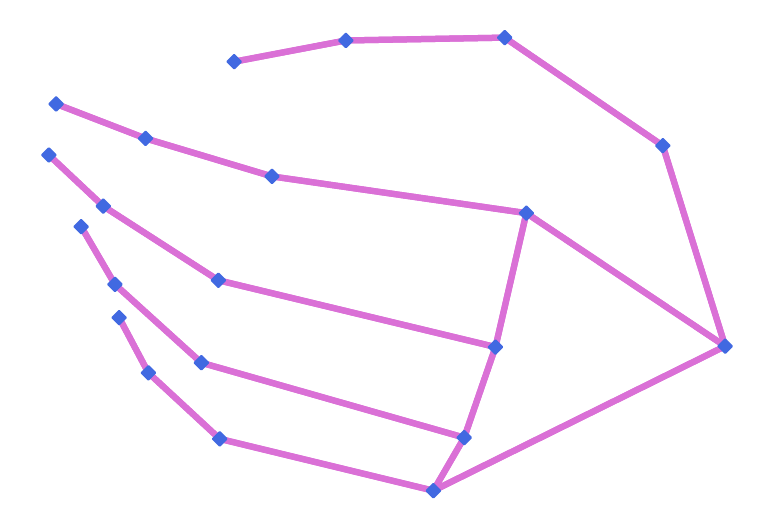}\\
      \includegraphics[width=\linewidth]{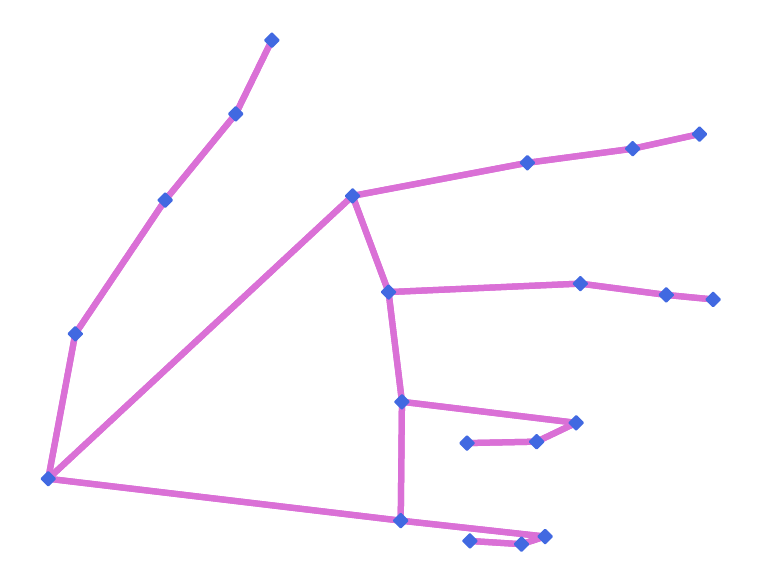}
    \end{subfigure}
    \caption{All extracted keypoints}
    \label{fig:individual_kp}
  \end{subfigure}
  \hspace{0.02\textwidth} % Small space between the subfigures
  \begin{subfigure}[m]{0.48\textwidth}
    \centering
    \includegraphics[width=\linewidth]{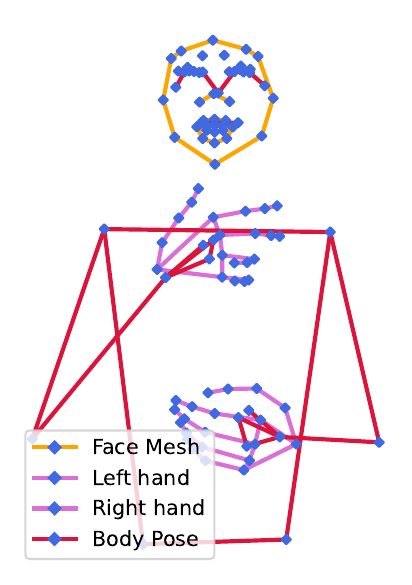}
    \caption{Keypoints used in our model}
    \label{fig:all_kp}
  \end{subfigure}
  \caption{We use only a subset of the keypoints extracted by MediaPipe. (a) shows all keypoints extracted by the individual MediaPipe models for the body, face, and hands. (b) shows the subset of keypoints that are used as input to our model.}
\end{figure}

For each hand, we obtain 21 keypoints. For body pose, we start with 33 keypoints but remove those corresponding to legs, as they are not essential for translation, leaving 25 keypoints. Lastly, we extract a dense face mesh containing 478 keypoints, from which we select 37 keypoints~\footnote{As defined in the YouTubeASL paper~\cite{yasl}.} that represent facial features. In total, we extract 104 keypoints. Figure~\ref{fig:individual_kp} shows the individual keypoints extracted for the body, face, and hands. Figure~\ref{fig:all_kp} illustrates the subset of keypoints used in our model.

\begin{figure}[hbpt!]
    \centering
    \begin{subfigure}[b]{0.59\textwidth}
        \centering
        \includegraphics[width=\textwidth]{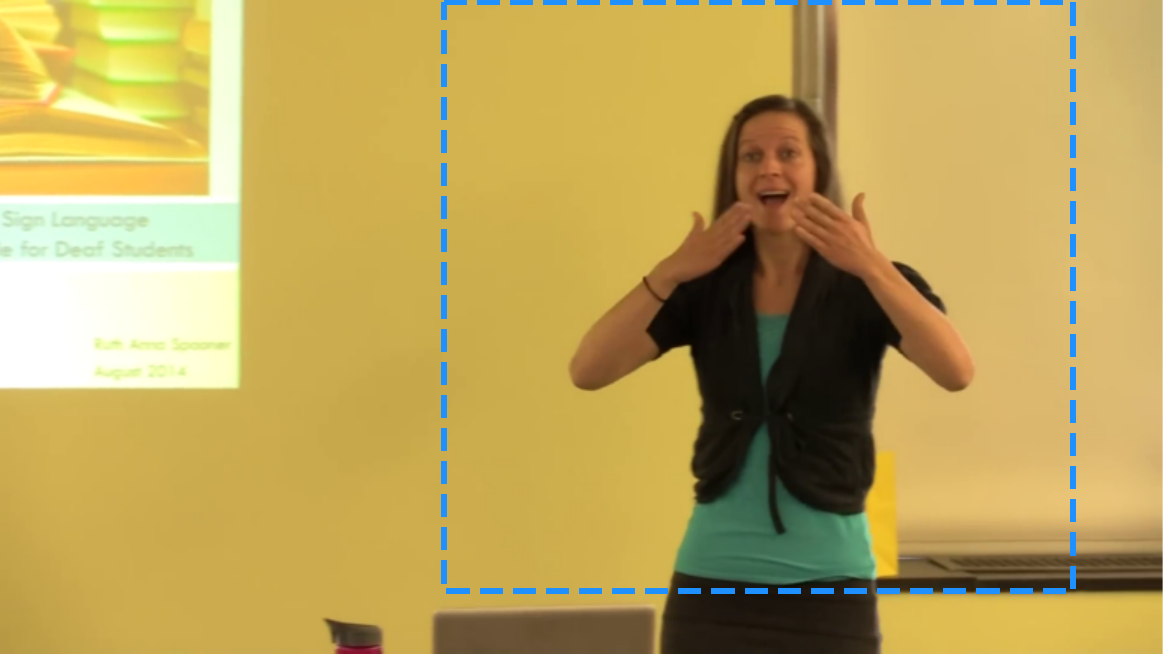}
        \caption{}
        \label{fig:}
    \end{subfigure}
    \begin{subfigure}[b]{0.332\textwidth}
        \centering
        \includegraphics[width=\textwidth]{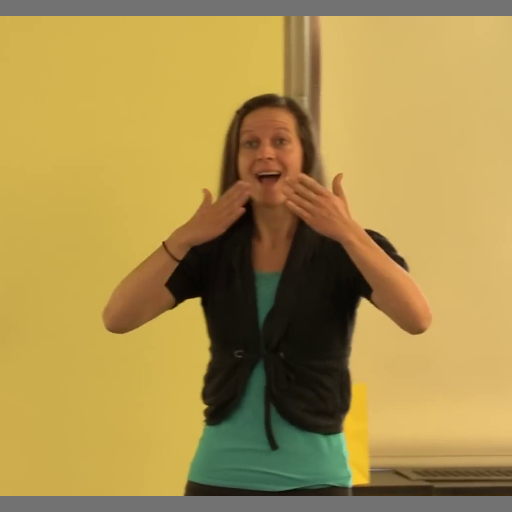}
        \caption{}
        \label{fig:}
    \end{subfigure}

    \caption{Video preprocessing based on sign space. (a) illustration of sign space in input frame, (b) cropped and padded frame.}
    \label{fig:video_preprocessing}
\end{figure}

Additionally, we apply normalization to all keypoints. For hand and face keypoints, we use local normalization, which involves creating a square bounding box around them to maintain the aspect ratio and then normalizing them to a range of -1 to 1. This provides a focused view of facial expressions and hand gestures. For body pose, we use global normalization, where all keypoints are normalized relative to the sign space, ensuring that all keypoints inside the sign space fall within the range of -1 to 1, see Fig.~\ref{fig:video_preprocessing}. Global pose normalization provides an overall view of the body pose and the relationships between different body parts.

Finally, we crop and pad frames to a square, resize them to a fixed resolution, and save them alongside the extracted keypoints. This preserves aspect ratios while removing unnecessary visual clutter, such as background. Simply resizing large videos to a smaller resolution without cropping the background could result in the loss of fine details in hand and face gestures. To reduce the sequence length, we remove every other frame, resulting in a preprocessed input consisting of 208-dimensional landmark vectors at half the original frame rate.

\subsection{Model}
Inspired by the YouTubeASL baseline approach, we used a similar, slightly modified version of the T5~\cite{T5} encoder-decoder transformer language model. Instead of the traditional approach of using a sequence of textual tokens as the input, we rather embed each 208-dimensional keypoint vector in the encoder using a single learnable linear layer. We experiment with three different T5 architectures: \mbox{T5-base}, \mbox{T5v1.1-base}\footnote{\url{https://github.com/google-research/text-to-text-transfer-transformer/blob/main/released_checkpoints.md}} for English and \mbox{mT5-base~\cite{xue2020mt5}} for both English and Arabic texts. 
\subsection{Training Pipeline} Our training pipeline follows a two-stage approach: pre-training on the YouTubeASL dataset and fine-tuning on the SSL dataset. This allows the model to first learn general sign language features from a large, diverse dataset (YouTubeASL) and then specialize on the target domain (SSL). The model was evaluated on three distinct test sets. 
% with signers \textit{xx} and \textit{xx} used for validation to ensure robustness and generalization.

\subsubsection{Pre-training} In the pre-training stage, the model was trained on the YouTubeASL dataset, a large-scale collection of SL videos paired with textual translations. This step enables the model to learn general SL features from a broad set of examples, which are crucial for transferring knowledge to the smaller SSL dataset. Since the original YouTubeASL paper~\cite{yasl} doesn't provide a training and validation split, we randomly sampled our own split with a ratio of 9:1 in such a way that the clips from the same source video can not be in both the training and validation subsets. We initialized the model with pre-trained T5-base weights, originally trained on textual data, and adapted it to process 208-dimensional keypoint vectors by embedding them into the encoder via a linear layer. The sequence-to-sequence framework was employed, where the encoder processed linearly mapped keypoint sequences and the decoder generated textual output. The results of the pre-trained models on the How2Sign dataset can be found in Table~\ref{tab: H2S results}.

\begin{table}[ht!]
    \centering
    \renewcommand{\arraystretch}{1.2}
    \setlength{\tabcolsep}{8pt} % Adjust column spacing for better alignment
    \begin{tabular}{lcccc}
        \hline
        \textbf{Model} & \textbf{BLEU-1} & \textbf{BLEU-2} & \textbf{BLEU-3} & \textbf{BLEU-4} \\
        \hline
        T5-Base & 22.21 & 3.91 & 1.49 & 0.62\\
        T5v1.1-Base & 24.52 & 4.54 & 1.71 & 0.72\\
        mT5-Base & \textbf{25.17} & \textbf{4.78} & \textbf{1.79} & \textbf{0.75}\\
        \hline
    \end{tabular}
    \vspace{5pt}
    \caption{How2Sign evaluation of our models pre-trained on YouTubeASL without any further fine-tuning on How2Sign.}
    \label{tab: H2S results}
\end{table}

\subsubsection{Fine-tuning} After pre-training, the model was fine-tuned on the SSL dataset to adapt to the specific characteristics of the English American Sign Language from YouTubeASL to the Arabic Saudi Sign Language. Fine-tuning helps bridge the gap between the general features of American and Saudi sign language. During this stage, the model was trained on keypoint-text pairs from the SSL dataset using the same sequence-to-sequence framework.

\subsubsection{Evaluation}
To evaluate the performance of our model, we examine its robustness and generalization across three distinct test scenarios, as outlined in Table~\ref{tab: data split}. For each testing scenario, we are providing the standard BLEU, BLEURT, and ROUGE-L scores.

\section{Experiments}

\subsection{Pre-training}
In the pre-training stage, we have trained T5-base, T5v1.1-base, and mT5-base models for a total of 200,000 training steps. The pre-training stage was conducted using 4 AMD MI250x GPU modules, split into 8 GCDs for each model. The T5-base model demonstrated efficient training with a learning rate of 0.001, while the T5v1.1-base and mT5-base models were trained with a smaller learning rate of 0.0004. For the pre-training, we use an effective batch size of 256 samples. Since mT5-base is a larger model, we use half the per-device batch size and double the gradient accumulation step to fully utilize our GPUs. We use Adafactor to optimize the model's parameters.

\subsection{Fine-tuning}
For fine-tuning, we conducted two rounds of experiments using three models trained on the English transcription: T5, T5v1.1, and mT5. The first round used the base model weights, and the second round used the weights pre-trained on YouTubeASL. The mT5 model was fine-tuned twice: once with the original Arabic transcription and once with translated English transcription using Google Translate, similar to T5 and T5v1.1. This resulted in a total of eight experiments.
All fine-tuning experiments were conducted on 8 NVIDIA A100-80GB GPUs. The learning rate was set to 0.001 with the AdamW optimizer and a linear LR scheduler, with a batch size of 16 per GPU (128 in total). For mT5, the batch size was reduced to 4, using gradient accumulation of 4 to mitigate memory issues. A weight decay of 0.01 was applied, and the models were trained for 100 epochs.

\subsection{Results}
The results of fine-tuning on the original base models' weights computed using SarceBLEU, BLEURT, and ROUGE-L metrics are shown in Table~\ref{tab:results1}, while the results for fine-tuning on YouTubeASL's checkpoints are presented in Table~\ref{tab:results2}. Since BLEURT is a trained metric and was not trained on Arabic, the results were not computed for the model fine-tuned on Arabic data. The tables demonstrate a clear increase in the scores across all test sets, supporting our claim that incorporating a large-scale sign language dataset during pre-training enhances the model's generalization across different and unseen languages. Notably, the results for Test-3 are significantly higher than those of the first two tests, as the sentences in the latter were unseen during training. Additionally, the consistent improvements across BLEU-1 to BLEU-4 and ROUGE-L indicate better word capture and phrase construction, noting the lower BLEU-4 scores compared to BLEU-1, as it focuses on longer phrases.

\begin{table*}[]
    \centering
    \scriptsize
    \renewcommand{\arraystretch}{1.0}
    \setlength{\tabcolsep}{5pt}
    \captionsetup{font=small,labelfont=bf}

    \begin{subtable}[t]{\textwidth}
        \centering
        \begin{tabular}{lcccccccc}
            \toprule
            \textbf{Model} & \textbf{BLEU-1} & \textbf{BLEU-2} & \textbf{BLEU-3} & \textbf{BLEU-4} & \textbf{BLEURT} & \textbf{ROUGE-L} \\
            \midrule
            T5-Base & 24.05 & \textbf{9.59} & \textbf{4.96} & \textbf{2.73} &-1.23 &22.1 \\
            T5v1.1-Base & \textbf{26.16} & 9.25 & 4.39 & 1.59 & -1.26 &\textbf{24.33} \\
            mT5-Base (Eng) & 23.63 & 8.32 & 3.98 & 1.46 & \textbf{-1.21} &21.36  \\
            mT5-Base (Ar) & 10.84 & 2.99 & 1.28 & 0.72 & - & 11.03 \\
            \bottomrule
        \end{tabular}
        \caption{Test-1: Unseen Signers – Unseen Sentences}
    \end{subtable}

    \vspace{0.5em}

    \begin{subtable}[t]{\textwidth}
        \centering
        \begin{tabular}{lcccccc}
            \toprule
            \textbf{Model} & \textbf{BLEU-1} & \textbf{BLEU-2} & \textbf{BLEU-3} & \textbf{BLEU-4} & \textbf{BLEURT} & \textbf{ROUGE-L} \\
            \midrule
            T5-Base & 24.46 & 9.14 & 4.53 & 2.01 & -1.24 & 22.22 \\
            T5v1.1-Base & \textbf{26.87} & \textbf{11.25} & \textbf{6.05} & 2.78 & \textbf{-1.21} & \textbf{25.17} \\
            mT5-Base (Eng) & 26.72 & 10.53 & 5.42 & \textbf{2.79} & -1.22 & 24.44\\
            mT5-Base (Ar) & 13.30 & 4.28 & 1.64 & 0.66 & - & 13.91\\
            \bottomrule
        \end{tabular}
        \caption{Test-2: Seen Signers – Unseen Sentences}
    \end{subtable}

    \vspace{0.5em}

    \begin{subtable}[t]{\textwidth}
        \centering
        \begin{tabular}{lcccccc}
            \toprule
            \textbf{Model} & \textbf{BLEU-1} & \textbf{BLEU-2} & \textbf{BLEU-3} & \textbf{BLEU-4}  & \textbf{BLEURT} & \textbf{ROUGE-L} \\
            \midrule
            T5-Base & 84.07 & 81.09 & 80.59 & 80.37 & 0.53 & 82.67\\
            T5v1.1-Base & \textbf{88.46} & \textbf{86.27} & \textbf{85.84} & \textbf{85.75} & \textbf{0.66} & \textbf{87.25} \\
            mT5-Base (Eng) & 87.76 & 85.62 & 85.22 & 85.16 & 0.64 & 86.71 \\
            mT5-Base (Ar) & 85.54 & 84.27 & 83.99 & 83.87 & - & 85.53\\
            \bottomrule
        \end{tabular}
        \caption{Test-3: Unseen Signers – Seen Sentences}
    \end{subtable}

    \caption{Relevant metrics for different T5 model variants across three test scenarios.}
    \label{tab:results1}
\end{table*}

\begin{table*}[]
    \centering
    \scriptsize
    \renewcommand{\arraystretch}{1.0}
    \setlength{\tabcolsep}{5pt}
    \captionsetup{font=small,labelfont=bf}

    % Subtable: Test-1
    \begin{subtable}[t]{\textwidth}
        \centering
        \begin{tabular}{lcccccc}
            \toprule
            \textbf{Model} & \textbf{BLEU-1} & \textbf{BLEU-2} & \textbf{BLEU-3} & \textbf{BLEU-4}  & \textbf{BLEURT} & \textbf{ROUGE-L} \\
            \midrule
            T5-Base & \textbf{35.89} & \textbf{17.33} & \textbf{11.14} & \textbf{7.48} & \textbf{-0.89} & \textbf{33.34} \\
            T5v1.1-Base & 34.76 & 16.79 & 10.05 & 5.56 & -0.98 & 31.5\\
            mT5-Base (Eng) & 33.50 & 16.07 & 9.77 & 5.66 & -1 & 30.72\\
            mT5-Base (Ar) & 16.75 & 5.40 & 1.86 & 0.81 & - & 16.79\\
            \bottomrule
        \end{tabular}
        \caption{Test-1: Unseen Signers – Unseen Sentences}
    \end{subtable}

    \vspace{0.5em}

    % Subtable: Test-2
    \begin{subtable}[t]{\textwidth}
        \centering
        \begin{tabular}{lcccccc}
            \toprule
            \textbf{Model} & \textbf{BLEU-1} & \textbf{BLEU-2} & \textbf{BLEU-3} & \textbf{BLEU-4}  & \textbf{BLEURT} & \textbf{ROUGE-L} \\
            \midrule
            T5-Base & \textbf{35.78} & \textbf{17.53} & \textbf{10.34} & \textbf{5.72} & \textbf{-0.89} & \textbf{33.15} \\
            T5v1.1-Base & 35.59 & 16.96 & 9.92 & 5.23 & -0.93 & 32.68\\
            mT5-Base (Eng) & 32.92 & 14.85 & 8.52 & 4.74 & -1.02 & 30.6\\
            mT5-Base (Ar) & 18.16 & 6.37 & 2.66 & 1.47 & - & 17.94 \\
            \bottomrule
        \end{tabular}
        \caption{Test-2: Seen Signers – Unseen Sentences}
    \end{subtable}

    \vspace{0.5em}

    % Subtable: Test-3
    \begin{subtable}[t]{\textwidth}
        \centering
        \begin{tabular}{lcccccc}
            \toprule
            \textbf{Model} & \textbf{BLEU-1} & \textbf{BLEU-2} & \textbf{BLEU-3} & \textbf{BLEU-4}  & \textbf{BLEURT} & \textbf{ROUGE-L} \\
            \midrule
            T5-Base & \textbf{95.17} & \textbf{94.02} & \textbf{93.78} & \textbf{93.67} &  \textbf{0.86} & \textbf{94.4}\\
            T5v1.1-Base & 94.50 & 93.16 & 92.83 & 92.58 & 0.85 & 93.97\\
            mT5-Base (Eng) & 94.64 & 93.44 & 93.19 & 93.04 & 0.85 & 94.09 \\
            mT5-Base (Ar) & 92.97 & 92.37 & 92.42 & 92.48 & - & 92.53\\
            \bottomrule
        \end{tabular}
        \caption{Test-3: Unseen Signers – Seen Sentences}
    \end{subtable}

    \caption{Relevant metrics of T5 model variants initialized with YouTubeASL pre-trained weights, evaluated on three test protocols.}
    \label{tab:results2}
\end{table*}

The results show that mT5 models trained on Arabic transcriptions perform poorly compared to those trained on English. This may be due to mT5 being trained on the mC4 dataset, where the representation of Arabic is smaller than that of English ~\cite{xue2020mt5}. Additionally, both tables reveal that mT5 models trained on English transcriptions consistently outperform those trained on Arabic. This suggests that translating non-English labels into English during mT5 training could improve model performance.

\section{Conclusion}
This paper tested the effectiveness of T5-based models for the Saudi Sign Language translation task using a novel dataset. In our experiments, we compared two main training protocols - direct training and pre-training on the large-scale American Sign Language (ASL) dataset. The SSL dataset incorporates three different testing protocols, which allowed us to systematically evaluate generalization across unseen signers and sentences. Additionally, the challenges posed by SSL-specific features, such as face coverings and unique grammatical structures, directly increase translation difficulty. Our results confirm the cross-linguistic transferability of sign language translation models and highlight the effectiveness of leveraging pre-training to overcome data scarcity issues in low-resource sign languages like SSL.

We would like to focus on two main research directions in our future work. Firstly, testing of different input modalities. In this paper, we utilized only pose as an input modality; for example, the DINO or MAE features can also encode relevant information. In fact, we were able to conduct some preliminary experiments with the DINO modality. However, we did not reach any satisfactory results with them. We argue that this can be caused by the fact that the dataset is relatively small, and therefore, models are not able to fully leverage the strength of the deep features.

Secondly, improvements in the preprocessing pipeline. In the current preprocessing pipeline, we entirely omit frames with multiple persons, resulting in less data for the training. Additionally, we would like to test different types of normalization, which seems to play a critical role in the quality of the preprocessing pipeline as demonstrated in~\cite{Spoter}.

\begin{credits}
\subsubsection{\ackname}  The authors with UWB affiliation have been supported by the grant of the University of West Bohemia, project No. SGS-2025-011. Computational resources were provided by the e-INFRA CZ project (ID:90254), supported by the Ministry of Education, Youth and Sports of the Czech Republic. We thank the Saudi Data \& AI Authority (SDAIA) for hosting the Winter School, where this work was conducted and for providing generous computing support.
\end{credits}

\bibliographystyle{splncs04}
\bibliography{mybib}

\begin{thebibliography}{10}
\providecommand{\url}[1]{\texttt{#1}}
\providecommand{\urlprefix}{URL }
\providecommand{\doi}[1]{https://doi.org/#1}

\bibitem{s24103112}
Al~Khuzayem, L., Shafi, S., Aljahdali, S., Alkhamesie, R., Alzamzami, O.: Efhamni: A deep learning-based saudi sign language recognition application. Sensors  \textbf{24}(10) (2024). \doi{10.3390/s24103112}, \url{https://www.mdpi.com/1424-8220/24/10/3112}

\bibitem{altamimi2023argument}
Altamimi, H.S., Alsager, H.N.: Argument structure and word order in saudi sign language. Journal of Language Teaching and Research  \textbf{14}(1),  203--214 (2023)

\bibitem{Spoter}
Boh\'a\v{c}ek, M., Hr\'uz, M.: Sign pose-based transformer for word-level sign language recognition. In: Proceedings of the IEEE/CVF Winter Conference on Applications of Computer Vision (WACV) Workshops. pp. 182--191 (January 2022)

\bibitem{PHOENIX-2014T}
Camgoz, N.C., Hadfield, S., Koller, O., Ney, H., Bowden, R.: Neural sign language translation. In: Proceedings of the IEEE Conference on Computer Vision and Pattern Recognition (CVPR) (June 2018)

\bibitem{chen2022simple}
Chen, Y., Wei, F., Sun, X., Wu, Z., Lin, S.: A simple multi-modality transfer learning baseline for sign language translation. In: Proceedings of the IEEE/CVF conference on computer vision and pattern recognition. pp. 5120--5130 (2022)

\bibitem{how2sign}
Duarte, A., Palaskar, S., Ventura, L., Ghadiyaram, D., DeHaan, K., Metze, F., Torres, J., Giro-i Nieto, X.: How2sign: a large-scale multimodal dataset for continuous american sign language. In: Proceedings of the IEEE/CVF conference on computer vision and pattern recognition. pp. 2735--2744 (2021)

\bibitem{llmsgoodsignlanguagetranslators}
Gong, J., Foo, L.G., He, Y., Rahmani, H., Liu, J.: Llms are good sign language translators. In: Proceedings of the IEEE/CVF Conference on Computer Vision and Pattern Recognition. pp. 18362--18372 (2024)

\bibitem{JWSign}
Gueuwou, S., Siake, S., Leong, C., Müller, M.: Jwsign: A highly multilingual corpus of bible translations for more diversity in sign language processing (2023), \url{https://arxiv.org/abs/2311.10174}

\bibitem{SignBERT+}
Hu, H., Zhao, W., Zhou, W., Li, H.: Signbert+: Hand-model-aware self-supervised pre-training for sign language understanding. IEEE Transactions on Pattern Analysis and Machine Intelligence  \textbf{45}(9),  11221--11239 (2023)

\bibitem{Jocher_Ultralytics_YOLO_2023}
Jocher, G., Chaurasia, A., Qiu, J.: {Ultralytics YOLO} (Jan 2023), \url{https://github.com/ultralytics/ultralytics}

\bibitem{ASLtoISL}
Kumar, M., Visagan, S.S., Mahajan, T.S., Natarajan, A.: Enhanced sign language translation between american sign language (asl) and indian sign language (isl) using llms (2024), \url{https://arxiv.org/abs/2411.12685}

\bibitem{LLaVASLT}
Liang, H., Huang, C., Xu, Y., Tang, C., Ye, W., Zhang, J., Chen, X., Yu, J., Xu, L.: Llava-slt: Visual language tuning for sign language translation (2024), \url{https://arxiv.org/abs/2412.16524}

\bibitem{lugaresi2019mediapipe}
Lugaresi, C., Tang, J., Nash, H., McClanahan, C., Uboweja, E., Hays, M., Zhang, F., Chang, C.L., Yong, M., Lee, J., et~al.: Mediapipe: A framework for perceiving and processing reality. In: Third workshop on computer vision for AR/VR at IEEE computer vision and pattern recognition (CVPR). vol.~2019 (2019)

\bibitem{T5}
Raffel, C., Shazeer, N., Roberts, A., Lee, K., Narang, S., Matena, M., Zhou, Y., Li, W., Liu, P.J.: Exploring the limits of transfer learning with a unified text-to-text transformer. Journal of machine learning research  \textbf{21}(140),  1--67 (2020)

\bibitem{SSVP-SLT}
Rust, P., Shi, B., Wang, S., Camgöz, N.C., Maillard, J.: Towards privacy-aware sign language translation at scale (2024), \url{https://arxiv.org/abs/2402.09611}

\bibitem{sprenger2012observations}
Sprenger, K., Mathur, G.: Observations on word order in saudi arabian sign language. Sign Language Studies  \textbf{13}(1),  122--134 (2012)

\bibitem{YouTube-SL-25}
Tanzer, G., Zhang, B.: Youtube-sl-25: A large-scale, open-domain multilingual sign language parallel corpus (2024), \url{https://arxiv.org/abs/2407.11144}

\bibitem{yasl}
Uthus, D., Tanzer, G., Georg, M.: Youtube-asl: A large-scale, open-domain american sign language-english parallel corpus. Advances in Neural Information Processing Systems  \textbf{36},  29029--29047 (2023)

\bibitem{Sign2GPT}
Wong, R., Camgoz, N.C., Bowden, R.: Sign2gpt: Leveraging large language models for gloss-free sign language translation (2024), \url{https://arxiv.org/abs/2405.04164}

\bibitem{xue2020mt5}
Xue, L.: mt5: A massively multilingual pre-trained text-to-text transformer. arXiv preprint arXiv:2010.11934  (2020)

\bibitem{yano2024multilingual}
Yano, C., Fukuchi, A., Fukasawa, S., Tachibana, H., Watanabe, Y.: Multilingual sentence-t5: Scalable sentence encoders for multilingual applications. arXiv preprint arXiv:2403.17528  (2024)

\bibitem{yin2023gloss}
Yin, A., Zhong, T., Tang, L., Jin, W., Jin, T., Zhao, Z.: Gloss attention for gloss-free sign language translation. In: Proceedings of the IEEE/CVF conference on computer vision and pattern recognition. pp. 2551--2562 (2023)

\bibitem{zhou2023gloss}
Zhou, B., Chen, Z., Clap{\'e}s, A., Wan, J., Liang, Y., Escalera, S., Lei, Z., Zhang, D.: Gloss-free sign language translation: Improving from visual-language pretraining. In: Proceedings of the IEEE/CVF International Conference on Computer Vision. pp. 20871--20881 (2023)

\bibitem{zhou2021improving}
Zhou, H., Zhou, W., Qi, W., Pu, J., Li, H.: Improving sign language translation with monolingual data by sign back-translation. In: Proceedings of the IEEE/CVF Conference on Computer Vision and Pattern Recognition. pp. 1316--1325 (2021)

\end{thebibliography}

% Only for Camera-ready:
% We publish our training scripts on our GitHub repository\footnote{\url{https://github.com/zeleznyt/T5_for_SLT/tree/main}}

\end{document}

% \maketitle
% \section*{Supplementary Material}
\begin{center}
    {\Large \textbf{Supplementary Material}}
\end{center}

\section{YASL Pre-training Results}
To provide additional context, we report the evaluation results on the How2Sign dataset in Table~\ref{tab: H2S results}. These results are only relevant for models pre-trained on the YouTubeASL dataset, as other models were specifically tuned for Saudi Sign Language and are not directly comparable in this setting.
\begin{table}[ht!]
    \centering
    \renewcommand{\arraystretch}{1.2}
    \setlength{\tabcolsep}{8pt} % Adjust column spacing for better alignment
    \begin{tabular}{c|cccc}
        \hline
        \textbf{Models} & BLEU-1 & BLEU-2 & BLEU-3 & BLEU-4 \\
        \hline
        \textbf{T5-Base} & 22.21 & 3.91 & 1.49 & 0.62\\
        \textbf{T5v1.1-Base} & 24.52 & 4.54 & 1.71 & 0.72\\
        \textbf{mT5-Base} & \textbf{25.17} & \textbf{4.78} & \textbf{1.79} & \textbf{0.75}\\
        \hline
    \end{tabular}
    \caption{How2Sign evaluation of our models pre-trained on YouTubeASL without any further fine-tuning on How2Sign.}
    \label{tab: H2S results}
\end{table}

\section{Training Hyperparameters}
The hyperparameters used in our experiments are listed in Table~\ref{tab:hyperparameters}. The values were mostly inspired by the original YouTubeASL paper, but have been adjusted to suit the specifics of our experiments. The seeds on the SSL-finetune columns contain two values, the first represents models trained on the base model weights, while the second corresponds to models trained on YouTubeASL checkpoints.
% \begin{table*}[t!]
%     \centering
%     \renewcommand{\arraystretch}{1.2}
%     \setlength{\tabcolsep}{4pt} % Adjust column spacing for better alignment
%     \begin{tabular}{l|ccc|cccc}
%         \hline
%         \textbf{} & \multicolumn{3}{c|}{\textbf{YouTubeASL Pre-training}} & \multicolumn{4}{c}{\textbf{SSL Fine-tuning}}\\
%         \textbf{Hyperparameter} & T5-Base & T5v1.1-Base & mT5-Base & T5-Base & T5v1.1-Base & mT5-Base \textsubscript{(en)} & mT5-Base \textsubscript{(ar)} \\
%         \hline
%         \textbf{Optimizer} & \multicolumn{3}{c|}{Adafactor} & \multicolumn{4}{c}{AdamW ~~~~ }\\
%         \textbf{Per device batch size} & 32 & ~~ 32 & 16 & 16 & 16 & 4 & 4\\
%         \textbf{Number of GPU devices} & \multicolumn{3}{c|}{8} & \multicolumn{4}{c}{8 ~~~~ }\\
%         \textbf{Gradient accumulation step} & 1 & ~~ 1 & 2 & 1 & 1 & 4 & 4 \\     
%         \textbf{Learning rate scheduler} & \multicolumn{3}{c|}{constant} & \multicolumn{4}{c}{linear ~~~~ } \\
%         \textbf{Weight decay} & \multicolumn{3}{c|}{0.0} & \multicolumn{4}{c}{0.01 ~~~~ } \\
%         \textbf{Base learning rate} & 0.001 & ~~ 0.0004 & 0.0004 & \multicolumn{4}{c}{0.001 ~~~~ } \\
%         \textbf{Training steps} & \multicolumn{3}{c|}{200,000} &  \multicolumn{4}{c}{18,800 ~~~~ }\\      
%         \textbf{Seed} & \multicolumn{3}{c|}{42} & 99, 42 &  0, 544 &  3037, 42 & 99, 3037 \\
%         \textbf{Max frame length} & \multicolumn{3}{c|}{250} & \multicolumn{4}{c}{600 ~~~~ } \\

%         \hline
%     \end{tabular}
%     \caption{Training hyperparameters.}
%     \label{tab: hyperparameters}
% \end{table*}

\begin{table*}[t!]
    \centering
    \renewcommand{\arraystretch}{1.2}
    \setlength{\tabcolsep}{6pt}
    \captionsetup{font=small,labelfont=bf}

    % Subtable 1: Pre-training
    \begin{subtable}[t]{0.9\textwidth}
        \centering
        \begin{tabular}{l|ccc}
            \toprule
            \textbf{Hyperparameter} & T5-Base & T5v1.1-Base & mT5-Base \\
            \midrule
            Optimizer & Adafactor & Adafactor & Adafactor \\
            Per device batch size & 32 & 32 & 16 \\
            Number of GPU devices & 8 & 8 & 8 \\
            Gradient accumulation step & 1 & 1 & 2 \\
            Learning rate scheduler & constant & constant & constant \\
            Weight decay & 0.0 & 0.0 & 0.0 \\
            Base learning rate & 0.001 & 0.0004 & 0.0004 \\
            Training steps & 200{,}000 & 200{,}000 & 200{,}000 \\
            Seed & 42 & 42 & 42 \\
            Max frame length & 250 & 250 & 250 \\
            \bottomrule
        \end{tabular}
        \caption{YouTubeASL Pre-training Hyperparameters}
    \end{subtable}

    \vspace{0.8em}

    % Subtable 2: Fine-tuning
    \begin{subtable}[t]{\textwidth}
        \centering
        \begin{tabular}{l|cccc}
            \toprule
            \textbf{Hyperparameter} & T5-Base & T5v1.1-Base & mT5-Base (en) & mT5-Base (ar) \\
            \midrule
            Optimizer & AdamW & AdamW & AdamW & AdamW \\
            Per device batch size & 16 & 16 & 4 & 4 \\
            Number of GPU devices & 8 & 8 & 8 & 8 \\
            Gradient accumulation step & 1 & 1 & 4 & 4 \\
            Learning rate scheduler & linear & linear & linear & linear \\
            Weight decay & 0.01 & 0.01 & 0.01 & 0.01 \\
            Base learning rate & 0.001 & 0.001 & 0.001 & 0.001 \\
            Training steps & 18{,}800 & 18{,}800 & 18{,}800 & 18{,}800 \\
            Seed & 99, 42 & 0, 544 & 3037, 42 & 99, 3037 \\
            Max frame length & 600 & 600 & 600 & 600 \\
            \bottomrule
        \end{tabular}
        \caption{SSL Fine-tuning Hyperparameters}
    \end{subtable}

    \caption{Training hyperparameters used for YouTubeASL pre-training and SSL fine-tuning.}
    \label{tab:hyperparameters}
\end{table*}

\section{Pose extraction}
\begin{figure}[htbp]
  \centering
  \begin{subfigure}[m]{0.255\textwidth}
    \centering
    \begin{subfigure}[m]{0.53\textwidth}
      \includegraphics[width=\linewidth]{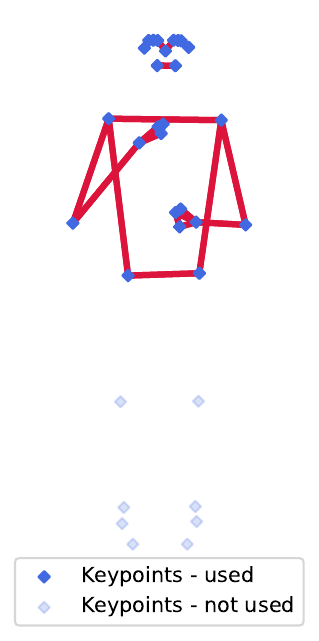}
    \end{subfigure}%
    \hfill
    \begin{subfigure}[m]{0.43\textwidth}
      \includegraphics[width=\textwidth]{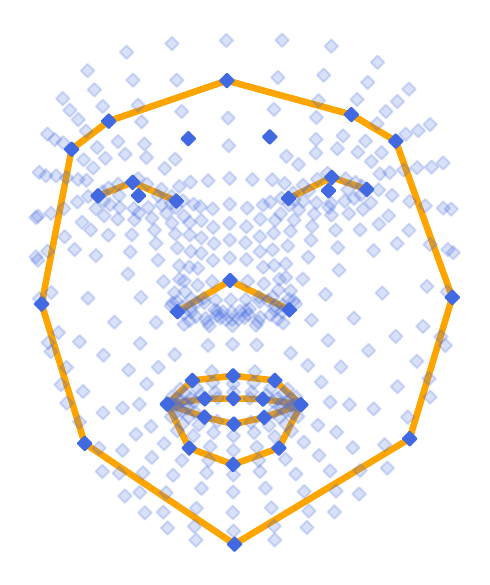}\\
      \includegraphics[width=\linewidth]{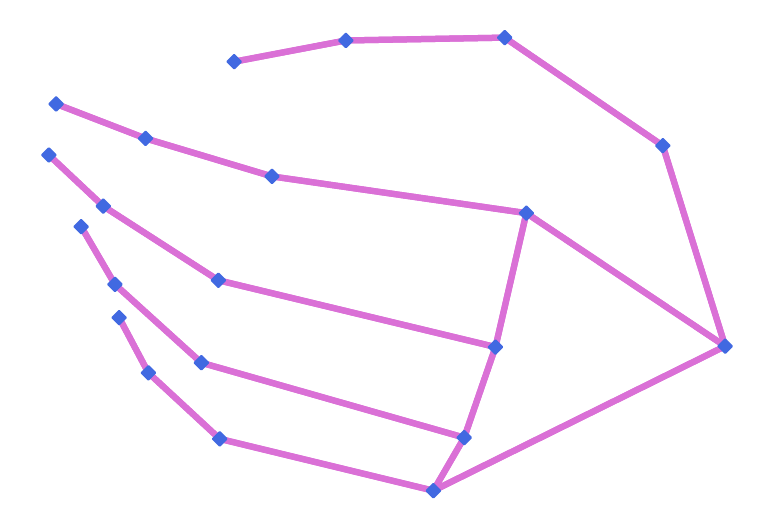}\\
      \includegraphics[width=\linewidth]{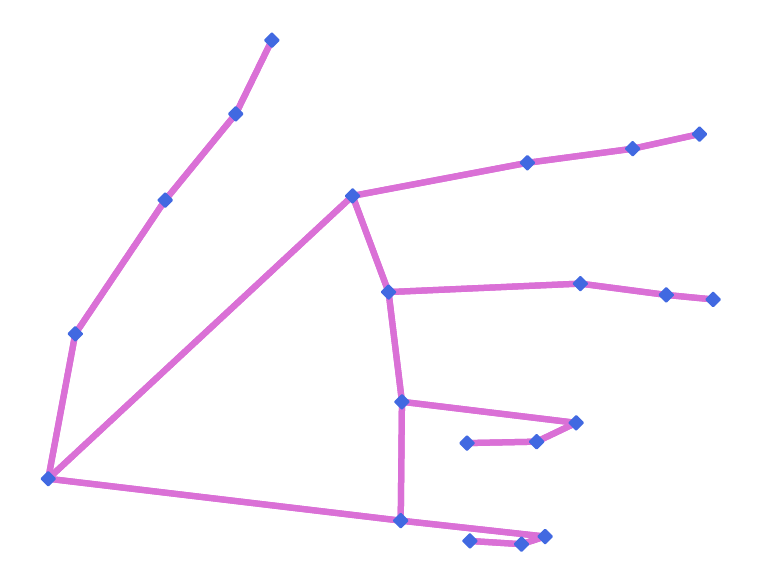}
    \end{subfigure}
    \caption{All extracted keypoints}
    \label{fig:individual_kp}
  \end{subfigure}
 \hfill
 \begin{subfigure}[m]{0.21\textwidth}
    \centering
    \includegraphics[width=\linewidth]{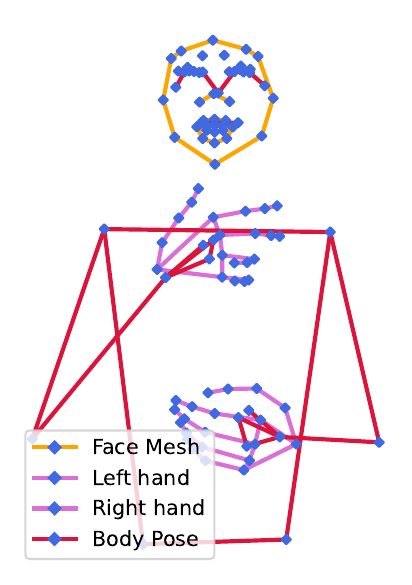}
    \caption{Keypoints used in our model}
    \label{fig:all_kp}
  \end{subfigure}
  \caption{We use only a subset of the keypoints extracted by MediaPipe. (a) shows all keypoints extracted by the individual MediaPipe models for the body, face, and hands. (b) shows the subset of keypoints that are used as input to our model.}
\end{figure}

We use MediaPipe to extract body pose, face mesh, and hand keypoints. MediaPipe provides 33 keypoints for body pose, 478 keypoints for the face mesh, and 21 keypoints for each hand. However, not all of these keypoints are necessary for sign language understanding and can therefore be removed.

The face mesh includes a large amount of redundant information. Following the YouTubeASL paper, we retain only 37 keypoints from the face mesh. Specifically, we keep keypoints at the following indices: 0, 4, 13, 14, 17, 33, 39, 46, 52, 55, 61, 64, 81, 93, 133, 151, 152, 159, 172, 178, 181, 263, 269, 276, 282, 285, 291, 294, 311, 323, 362, 386, 397, 402, 405, 468, 473.

For the body pose, we discard the lower-body keypoints, as they are generally irrelevant to sign language. Specifically, we remove all body pose keypoints with indices $\geq$ 25. All hand keypoints are retained in full.

In total, we extract 104 keypoints. Figure~\ref{fig:individual_kp} shows the individual keypoints extracted for the body, face, and hands. Figure~\ref{fig:all_kp} illustrates the subset of keypoints used in our model.